# DEMorphy, German Language Morphological Analyzer

Duygu Altinok

Berlin, Germany

**Abstract.** DEMorphy is a morphological analyzer for German. It is built onto large, compactified lexicons from German Morphological Dictionary. A guesser based on German declension suffixes is also provided. For German, we provided a state-of-art morphological analyzer. DEMorphy is implemented in Python with ease of usability and accompanying documentation. The package is suitable for both academic and commercial purposes with a permissive licence.

1. INTRODUCTION

Morphological analysis is analysis of structure of the word forms. For morphologically complex languages such as Turkish, Finnish, German; a full morphological analysis provides information about word category (noun, verb, adjective etc.), word lemma, number, gender, person etc. In these languages morphology derives syntax, word analysis also provides information about possible POS tags (or vice versa, syntax determines feasible morphological forms. Here, morphology↔syntax implication is always two-sided). Morphological analysis units, either rule-based or unsupervised statistical, is an important step of modern NLP pipelines. From statistical machine translation to sentiment analysis, training morphology aware NNs to semantic search; morphological analysis keeps its importance in modern NLP pipelines.

Morphological generation is the process of creating word forms as appear in the language from a base form and analysis. Roughly, given the lemma(s) and a list of inflections and derivations, finding the correct word form is called generation.

DEMorphy is a morphological analyzer for German language. The package is available[1] under MIT Licence and itself uses another open source library.

DEMorphy is an efficient, DAFSA based Python implementation. Unlike other similar freely available German analyzers SMOR[2] and Morphisto[3], DEMorphy is implemented in native Python and does not require any extra Python bindings or extra system calls at runtime. DEMorphy is production-ready, brings predictable runtime quality and provides predictable concurrency behaviour (following well-understood Python concurrency model).

Rest of the paper is organized as follows: Section 2 explains details about the German Morphological Dictionary[4]. In section 3, software architecture and implementation details will be given. Section 4 contains details of the morphological analysis process. Section 5 outlines a roadmap for future improvements and new features.

---

[1] https://github.com/DuyguA/DEMorphy
[2] http://www.cis.uni-muenchen.de/~schmid/tools/SMOR/
[3] http://www1.ids-mannheim.de/lexik/home/lexikprojekte/lexiktextgrid/morphisto.html
[4] https://github.com/DuyguA/german-morph-dictionaries

## 2. GERMAN MORPHOLOGICAL DICTIONARY

gmd[5] is a German lexicon and analysis dictionary, it contains lexicon words and list of their all possible analysis. gmd is generated by running our in-house morphological analyzer on Wikidumps[6] corpus. Interjections, modular verbs, auxiliary verbs, particles, conjunctions, determiners and articles (i.e. closed grammatical categories of the German language) is processed with extra care and appears at the beginning of the text file. Dictionary format is plain text, which is suitable for converting to other formats such as XML, for different purposes. gmd consists of

(a) Word forms as they appear in written language, inflected, derived forms and compounds included. Each word form is followed by list of its possible analysis.
(b) Experimental analysis dictionary
(c) List of all lemmas that occur in (a)
(d) List of all paradigms that occurs in (a)

A typical entry looks like:

```
gegangen
gegangen ADJ,pos,<pred>
gegangen ADJ,pos,<adv>
gehen V,ppast
```

Analysis lines are of the form (lemma, paradigm). A paradigm is a list of
- Grammatical category as first entry, always
- Gender, number, person, positive/comparative/superlative etc. list of inflections

separated by commas.

Our in-house analyzer can
- split compounds i.e. list all possible splits of a compound,
- show bounding morphemes between compound elements,
- analyze derivation,
- analyze inflection.

Also, our in-house analyzer is
- FST based and compiled from marked lexicon,
- compiled with OpenFST,
- efficient, can be processed by Python via PyFST directly without needing external Python binders; fully Python compatible and reliable for software production.

---

[5] https://github.com/DuyguA/german-morph-dictionaries
[6] https://dumps.wikimedia.org/dewiki/latest/

However, DEMorphy does not include all capabilities of our in-house analyzer. We currently provide only the inflection analysis. Derivational analysis and compound splitting is not included. Hence a typical compound analysis looks like:

Rohrohrzucker
Rohrohrzucker NN,masc,acc,plu
Rohrohrzucker NN,masc,acc,sing
Rohrohrzucker NN,masc,nom,sing
Rohrohrzucker NN,masc,gen,plu
Rohrohrzucker NN,masc,dat,sing
Rohrohrzucker NN,masc,nom,plu

whereas output of our in-house tool is:

Roh<#>rohr<#>zucker NN,masc,acc,plu
Roh<#>rohr<#>zucker NN,masc,acc,sing
Roh<#>rohr<#>zucker NN,masc,nom,sing
Roh<#>rohr<#>zucker NN,masc,gen,plu
Roh<#>rohr<#>zucker NN,masc,dat,sing
Roh<#>rohr<#>zucker NN,masc,nom,plu.

While preparing analysis of the compounds, first we took all possible splits, then filtered by (1) feasible POS tag combinations (2) with a language model to eliminate "nonsense" analyses. For instance, in the above example analysis rohr<#>ohr<#>zucker (*pipe ear sugar*) is eliminated by (2).

The experimental word forms is the list of word analysis that is produced by our "guesser". Basically we separated analysis that developers should use at their "own risk" into a separate list. The dictionary size is 340MB, experimental forms dictionary is sized 15MB. An encoded form is also available, lemma and paradigm strings are encoded to numbers pointing to the lemmas list and the paradigms list. The encoded dictionary size is 135MB, list of all lemmas is 18MB and paradigms list is 20KB. Number of all inflectional paradigms is 643 with a total number of 1.187.013 possible lemmas. Together with the experimental forms the morphological dictionary contains 12.066.971 entries with a lexicon size 2.168.203 word forms.

## 3. SOFTWARE ARCHITECTURE

DEMorphy is implemented as a Python library. Both Python 2.x and Python 3.x are supported. Documentation is available online.

The average parsing speed is around 15 000 - 20 000 words per second. Memory consumption of the compacted dictionaries is about 100 MB, together with the Python interpreter 120 MB. Users are provided with library code for obtaining word analysis, word lemma, possible POS tags in both Stuttgart[7] (STTS) and Penn Tree Bank[8] (PTB) tags as well as iterators over the all lexicon. All possible analysis of a given word is provided, choosing which one to use is to be determined the user.

Cache implementations are also provided within the package. An LRU cache implementation provides users obtaining analysis with cache lookup. The LRU cache proves efficiency on average length German text, conjunctions, pronouns, auxiliary verbs, modal verbs, articles, determiners and context words appear over and over again. Instead of computing same word's analysis over and over, user can cache the analysis result. An unlimited cache implementation is also provided, however not recommended; LRU cache would be enough and well-performing.

## 4. ANALYSIS

DEMorphy relies on a compacted form of German Morphological Dictionary as we remarked before. End users do not have to compile the compacted dictionary themselves, we will deliver precompiled dictionaries on updates.

Morphological analysis indeed is just dictionary lookup in DAFSA. Given a word, we fetch all possible analysis strings. From the plain text dictionary file, one can build a Python dictionary (a hashmap) and query the input words. However, this approach might have some problems:

- Memory-killer: Words belong to same paradigms usually have same endings, for instance *imfendem*, *informierendem*, *gehendem*, *abgehendem*, *umgehendem;* especially declensions. Also, inflections of the same lemma share long prefixes, for instance *kurieren*, *kurierend*, *kurierende*, *kurierender*, *kurierendes*, *kurierst*. Common prefixes such as *um*, *ab* also occurs frequently e.g. *melden*, *anmelden*, *abmelden*, *nachmelden*, *ummelden*, *vermelden*, *weitermelden*, *zurückmelden*. It is not very storage friendly to store the same substrings again and again.
- Time efficiency: Hashmap provides $\Theta(1)$ lookup, however hash function needs to go over all characters of the input string 1-by-1. If one is interested in fuzzy lookup, then all options need to be generated and looked up 1-by-1. For instance, if "u" can be both "u" and "ü", then we need 2 lookups. This situation happens quite a lot with the German umlauts.

---

[7] http://www.ims.uni-stuttgart.de/forschung/ressourcen/lexika/TagSets/stts-table.html
[8] https://www.ling.upenn.edu/courses/Fall_2003/ling001/penn_treebank_pos.html

To compactify the common substrings and enable fast access together with fast fuzzy lookup; DEMorphy stores the words in a DAFSA using DAWG[9] library. DAWG is a Cython based, low-level, efficient, unicode supporting Python library. Information about lemma and paradigm string is encoded as positive integers, pointing to real strings in lemma list and paradigm list.

**DAFSA** Directed acyclic graphs are widely used in NLP, computational morphology and IR tasks. DEMorphy stores word forms in DAFSA as well, due to
- Memory efficiency
- Fast lookup
- Fast fuzzy lookup
- Flexible iteration support

Here is an example how common endings are stored:

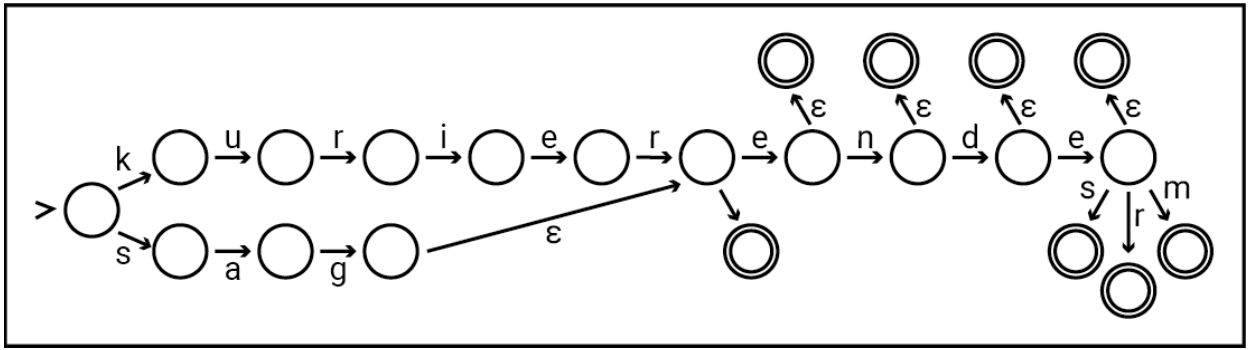

Figure 1: Common Endings Storage

DAFSA for German lexicon of 12.066.971 entries is about 100MB, whereas the plain text file is 135MB.

Fuzziness caused by the German umlaut characters are handled efficiently by DAFSA as well. Take the example "grün". It is quite possible that it is written as "grun" by a non-German keyboard. We provide DAFSA a character mapping, in which characters might represent more than one character, a character set. DAWG allows 1-to-1 character mappings, hence we provided u → ü, o → ö, a → ä mappings.

While scanning the input string "grun", at the second state, DAFSA permits possible u → ü replacement and correctly reaches the acceptance state. DAWG allows only 1-to-1 mappings, hence we dealt with ue → ü , oe → ö, ae →  ä, ss → ß,  ß → ss with another method.

---

[9] https://pypi.python.org/pypi/DAWG

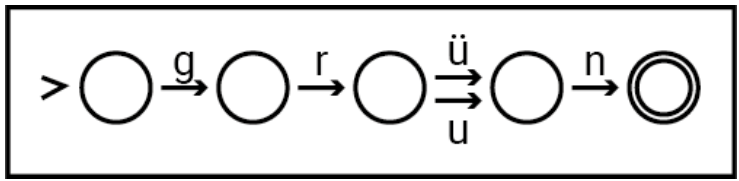

Figure 2: Input String "grun" scanning left to right

**OOV** OOV is handled by prefix and suffix analogies. Currently we support obvious verb endings such as *test*, *tet*, *tem*, *ten* etc. by a suffix-prefix-infix analyzer. In the future, we plan to expand recognition by a character level language model.

**Compound Words** As we remarked before, DEMorphy does not carry all capabilities of our in-house analyzer. Our in-house analyzer analyzes compounds completely, exhibiting component word boundaries, linking morphemes and then the derivations and the inflections. DEMorphy does not provide a full analysis, rather provides the lemma and the inflections; just as rest of the lexicon words. However, compound words processed carefully before shipped to the morphological dictionary. Our in-house tool generated all possible splits, then we filtered by impossible POS tag combinations (e.g. beiden is not bei<#>den) and a language model. We already gave the example "Rohrohrzucker", has 2 possible splits rohr<#>ohr<#>zucker *(pipe ear sugar)* and roh<#>rohr<#>zucker (*raw cane sugar*). The language model eliminated the first form because "it does not make sense", i.e. it admits a very low probability. In general, we used the heuristic that "less split is better than more splits". If a compound admits a 2 words split and a 3 words split, we preferred the former.

**Words with a Hyphen** is processed by STTS notation, *TRUNC*. We included them into lemma, rather than the analysis. For instance, lemma of "U-Bahn" is *U-(TRUNC)Bahn* and lemma of "U-Bahn-Station" is *U-(TRUNC)Bahn-(TRUNC)Station*.

**Other Types of Tokens** For the tokens that are part of the written language but not German language lexicon, for instance e-mails, date strings, url strings etc. DEMorphy contains a special processing unit. This unit first evaluates if word belongs to one of these classes, if so do not ask the analyzer and directly provide the token type as a result.

5. FUTURE WORK

New versions will include
- Better analogy analyzer
- Character level language model support
- Detailed support on the geographical names, proper nouns (first names, last names, company names, brand names), abbreviations
- More work on the experimental dictionary

Though DEMorphy is fast enough, there is always room for further time efficiency improvements.

## 6. REFERENCES

Python Software Foundation. Python Language Reference, version 2.7. Available at
http://www.python.org

Escartín, C. P. (2014). Chasing the Perfect Splitter: A Comparison of Different Compound Splitting Tools. In Proceedings of the Ninth International Conference on Language Resources and Evaluation, pages 3340–3347, Reykjavik

Fabienne Fritzinger , Alexander Fraser, How to avoid burning ducks: combining linguistic analysis and corpus statistics for German compound processing, Proceedings of the Joint Fifth Workshop on Statistical Machine Translation and MetricsMATR, p.224-234, July 15-16, 2010, Uppsala, Sweden

Helmut Schmid, Arne Fitschen and Ulrich Heid: SMOR: A German Computational Morphology Covering Derivation, Composition, and Inflection, *Proceedings of the IVth International Conference on Language Resources and Evaluation (LREC 2004)*, p. 1263-1266, Lisbon, Portugal

Koehn, P., Knight, K.: Empirical Methods for Compound Splitting. Proc. 10th Conf. of the European Chapter of the Association for Computational Linguistics (EACL). Budapest, Hungary (2003) 347–35

Korobov, Mikhail: Morphological Analyzer and Generator for Russian and Ukranian Languages, CoRR, abs/1503.07283, 2015

Zielinski A., Simon C., Wittl T. (2009) Morphisto: Service-Oriented Open Source Morphology for German. In: Mahlow C., Piotrowski M. (eds) State of the Art in Computational Morphology. SFCM 2009. Communications in Computer and Information Science, vol 41. Springer, Berlin, Heidelberg